\documentclass[12pt]{article}

\usepackage[margin=1in]{geometry}
\usepackage{amsmath, amssymb, amsthm}
\usepackage{graphicx}
\usepackage{caption}
\usepackage{subcaption}
\usepackage{booktabs}
\usepackage{hyperref}
\usepackage{cleveref}
\usepackage{enumitem}
\usepackage[utf8]{inputenc}
\usepackage{csquotes}
\usepackage{float}
\usepackage{xcolor}
\DeclareUnicodeCharacter{2264}{\ensuremath{\leq}}
\DeclareUnicodeCharacter{2265}{\ensuremath{\geq}}
\usepackage{multirow}
\usepackage[superscript]{cite}

\theoremstyle{definition}

\theoremstyle{plain}

\theoremstyle{remark}

\title{\textbf{Optimization and Constraint Modeling using LLMs with a Retrieval Augmented Generation Process}}
\author{
Prateek Roy \\
\small Mentor: Akash Singirikonda
}
\date{November 2025}

\begin{document}

\maketitle

\begin{abstract}
Both optimization modeling and constraint modeling are non-trivial problems requiring deep domain expertise along with proficiency in modeling formalism languages. Despite their importance across logistics, healthcare, and supply chain management, a critical gap exists: current large language models (LLMs) regularly produce structurally inconsistent or incomplete optimization formulations, particularly in combinatorial settings, and no scalable retrieval-based remedy has been systematically evaluated. This paper addresses that gap by evaluating whether a Retrieval-Augmented Generation (RAG) pipeline built on a curated synthetic dataset can meaningfully improve LLM optimization modeling performance.

A total of 500 optimization problems were synthesized using seed descriptions obtained from the Text2Zinc dataset and professional personas created using an LLM. These problems were specified using JSON structure and associated with validated Python solver scripts that could solve them. The optimization problems were then encoded in a vector database of Chroma type. Once an instance of an inference problem was provided, semantically similar problems were retrieved, processed by a semantic gateway and used as contextual instances to provide guidance to the LangChain LLM agent.

Results: Three benchmark testbeds were used to conduct evaluation tests for the proposed RAG augmented pipeline under the Qwen 3 30B Instruct model. The performance of accuracy rose from 40\% to 72\% on NL4OPT, 40\% to 56\% on MAMO Easy, and 32\% to 56\% on MAMO Complex.

Conclusions: The use of optimization problems generated using semantically validated examples greatly improves both solution accuracy and structure. The findings point out that the combination of synthetic dataset generation with RAG provides an effective and accessible alternative approach compared to fine-tuning. These results suggest that domain-specific synthetic corpora paired with retrieval augmentation can serve as a practical pathway for deploying LLM-based optimization tools in real-world decision-support contexts without requiring costly model retraining.

\end{abstract}

\section{Introduction}

The mathematical modeling of complex problems by imposing constraints and using optimization techniques is one of the core areas in decision support systems across logistics, healthcare, energy, transportation, and supply chain management. Due to an increase in problem size and complexity, large language models have been increasingly used in optimization problems \cite{llm_optimization2023}, yet state-of-the-art LLMs struggle to produce accurate models when tasks require strict structure or multi-step reasoning \cite{llm_optimization2023}. This paper focuses on improving the accuracy of optimization models generated by an LLM, as greater accuracy enables more efficient decision-making and resource utilization.

The contribution of this work is to enhance the relationship between natural language problem statements and formal optimization models, thus contributing to the development of explainable AI models. By showing that retrieval-augmented generation can greatly improve LLM performance on structured modeling problems, this study paves the way for developing a transparent tool capable of assisting field experts without optimization model language proficiency.

The solution proposed for this issue is based on building a vector database that will contain systematic and formatted data regarding constraint and optimization problems and their solutions. Each problem contains a description written in the form of a natural language statement, a formal description of the problem in the JSON format, and its Python solution using a model library. All datasets are formalized in order to minimize confusion and facilitate quick search. Using the Retrieval-Augmented Generation (RAG) architecture \cite{rag2020,realm2020,fusion_in_decoder2020}, it is possible to extract semantically similar examples from the vector database, which can serve as in-context cues for an LLM in order to make it produce appropriate solutions.

We make three major contributions to the current state of research:

\begin{itemize}

    \item \textbf{Construction of Dataset:} We created a systematically designed dataset of more than 500 artificial constraint/optimization problems, every one of which has a verified solution written in Python code using a modeling library after being prompted by GPT-5.
    
    \item \textbf{Vector Database for RAG:} In this regard, we have implemented a vector database system for RAG that helps an LLM retrieve semantically significant examples while generating outputs.
    
    \item \textbf{Accuracy Improvement Pipeline:} We propose and experiment with a Retrieval-Augmented Generation (RAG) modeling pipeline and show the effectiveness of the pipeline at improving the accuracy of constraint and optimization problem formulations generated by the LLM.

\end{itemize}

Research Objectives. This study pursues three concrete objectives: (1) to construct a diverse, validated synthetic dataset of constraint and optimization problem-solution pairs suitable for use as a retrieval corpus; (2) to design and implement a RAG-based inference pipeline that retrieves semantically similar solved problems and uses them to guide LLM output; and (3) to empirically measure the accuracy improvement conferred by RAG augmentation relative to a direct LLM baseline across multiple publicly available benchmarks.

Constraint satisfaction and optimization problems formulated in LP, MILP, MINLP, and CP can be investigated using this approach. This will be tested using three publicly available benchmark problems, using one base model (Qwen 3 30B Instruct), and 500 generated instances. To apply the results beyond this particular type of models, solver backends, and types of constraint satisfaction and optimization problems, further testing needs to be performed. The data are artificially created. Solver scripts have been checked for compilation and execution, but correctness of all optimality certificates provided cannot be independently confirmed by domain experts.

\medskip

This work is grounded in the Retrieval-Augmented Generation framework introduced by Lewis and colleagues \cite{rag2020}, which posits that non-parametric retrieval mechanisms can complement the parametric knowledge of autoregressive language models to reduce hallucination and improve factual grounding. Applied to mathematical programming, this framework rests on a key assumption: that semantic similarity in natural language problem descriptions implies structural similarity in the underlying optimization formulations. Under this lens, a curated vector database of solved problems functions as an external knowledge store, and cosine similarity in embedding space serves as the retrieval signal. The semantic gateway introduced in this paper extends the framework by adding a conditional acceptance layer, drawing on the adaptive retrieval theory of Jeong, Baek, Cho, Hwang, and Park \cite{adaptiverag2024}, which shows that selectively including retrieved context outperforms unconditional injection.

\medskip

The research consists of three phases, namely: (1) synthetic data creation through LLM persona prompts and problem synthesis using JSON format; (2) embedding and indexing of problem-solution pairs in a Chroma vector database; and (3) RAG-based inference through an intelligent LangChain agent with semantic filtering of retrieved context by cosine similarity. Each phase is carefully designed to be modular: the dataset construction pipeline (Section~3), data cleaning (Section~4), and exploratory analysis (Section~5) are prerequisites to the retrieval and inference pipeline detailed in Section~6. Evaluation against three benchmarks is reported in Section~7 and interpreted in Section~8.

\section{Background and Related Work}

\subsection{Optimization and Constraint Programming Overview}

Optimization methods and Constraint Programming provide the mathematical basis to tackle complicated decision-making tasks in areas like logistics, manufacturing, energy models, health, and transport. Some widely used paradigms in modeling these complex decision tasks include Linear Programming (LP), Mixed-Integer Linear Programming (MILP), Constraint Programming (CP), and Mixed-Integer Nonlinear Programming (MINLP).

A problem can be represented in terms of linear objectives and constraints in both LP and MILP, with the added ability in MILP for integer or binary variables. The approach in CP focuses on search with guidance from constraint satisfaction, rather than linear constraints. The addition in MINLP of nonlinear terms and mixed integrality significantly increases difficulty in computation compared with the constrained format.

However, such mathematical models require domain knowledge. In order to become effective models, the following are needed:

\begin{itemize}
\item \textbf{Precise definitions of variables}, including correct domains and relationships

\item \textbf{Complete and logically consistent constraints}, ensuring feasibility and structural correctness

\item \textbf{An appropriately structured objective}, formulating cost minimization, utility maximization, or the efficiency of resources

\end{itemize}

Although small inaccuracies in a model formulation may result in a model being unfeasible, suboptimal, or inconsistent, these issues trigger interest in an area of AI-assisted modeling, whereby language models may be useful in speeding up model formulation and enabling non-experts to perform formulations.

\subsection{Language Models for Mathematical Reasoning}

Large Language Models (LLMs) such as GPT-3/4, LLaMA, Qwen, and Mistral have been making great strides in natural language understanding and code synthesis and reasoning. It was observed that the LLMs could generate algebraic proofs and code and could also handle solvers and symbolic engines. Still, mathematical proof and constraint reasoning pose a great challenge \cite{llm_optimization2023}.

Research work has pointed to several limitations of LLMs in optimization modeling. Existing work highlights several weaknesses of LLMs in optimization modeling:

\begin{itemize}
\item \textbf{Logical inconsistency}

\item \textbf{Specification errors} -- There could be misinterpretations of objectives, inverted guidance (min vs max), or unserved constraints for the scenario.

\item \textbf{Incomplete formulations} -- Variables may be absent, constraints may be ignored, or critical parameters may not be defined.

\item \textbf{Tendency to hallucinate mathematical structure.} The model can add fake variables or parameters to equations when there is no mathematical structure to model from.

\end{itemize}

Since optimization problems necessarily involve some degree of \textit{structured, internally consistent output} -- typically given in terms of JSON schemas, algebraic expressions, and solver-ready Python code -- LLMs are prone to falling short of the desired degree of precision when using just internal knowledge.

These constraints push for the integration of retrieval systems, which provide LLMs with access to elaborate example formulations.

\subsection{Retrieval-Augmented Generation (RAG)}

Retrieval-Augmented Generation (RAG) is a combination of neural text generation and a process of external knowledge retrievals \cite{rag2020}. Before the LLM produces an output, it queries a vector database to yield the most semantically relevant documents or examples. Then, it uses these documents or examples to ground its generation process. Thus, it cuts down on hallucinations and keeps it structured.

RAG has attained excellent results in several domains \cite{rag2020,realm2020,fusion_in_decoder2020}:

\begin{itemize}

\item \textbf{Code generation:} Offering useful code snippets or APIs significantly lowers syntax errors and phantom calls to libraries.
\item \textbf{Question answering}: RAG enhances factual grounding by requiring models to reference their retrieved evidence.
\item \textbf{Scientific and mathematical reasoning:} Examples drawn from specific corpora enhance step-by-step reasonings and ensure that wrong deductions are not made.

\end{itemize}

These successes indicate the promise of RAG for optimization modeling, where a RAG-powered LLM can reuse proven formulation templates, avoid common pitfalls such as missing constraints or improper domains, and produce solver-ready code more reliably than zero-shot inference alone.

\subsection{Datasets for Optimization \& CP Modeling}

Several public datasets provide examples of optimization and constraint problems, but none offers the scale, consistency, or structure needed for high-quality LLM training \cite{text2zinc2022,nl4opt2022}.
 
\begin{itemize}

\item \textbf{Text2Zinc} \cite{text2zinc2022} comes with natural language descriptions, together with CP-SAT (MiniZinc) formulation. But the problems are too diverse, and the set of problems is limited.

\item \textbf{NL4OPT} \cite{nl4opt2022} provides natural language optimization problems related to LP/MILP translation. However, diversity in problems and solution details are not fully explored.

\item \textbf{OptMATH}-style and mathematical reasoning corpora deal more with mathematical expression manipulation and less with complete MILP/CP formulation tasks.

\end{itemize}

The fact that large, consistent datasets containing \textit{paired} natural language descriptions and \textit{validated solver code} are scarce hinders progress in the LLM-based optimization modeling field, motivating the construction of a synthetic dataset offering well-structured JSON/solver pairs, persona-driven language variation \cite{persona_ai2025}, and ground-truth correctness verified by solver execution.

\section{Dataset Construction}
\label{sec:dataset}

We will go over the creation of a synthesis dataset comprising optimization and models with solver code. It will entail four steps: (i) extraction of basic problem descriptions from a given set of models, (ii) creation of professional personas that fit the context, (iii) creation of JSON-formatted models that model these optimizations, and (iv) generation of full Python code for the models.

\begin{figure}[htbp]
    \centering
    \includegraphics[width=1\linewidth]{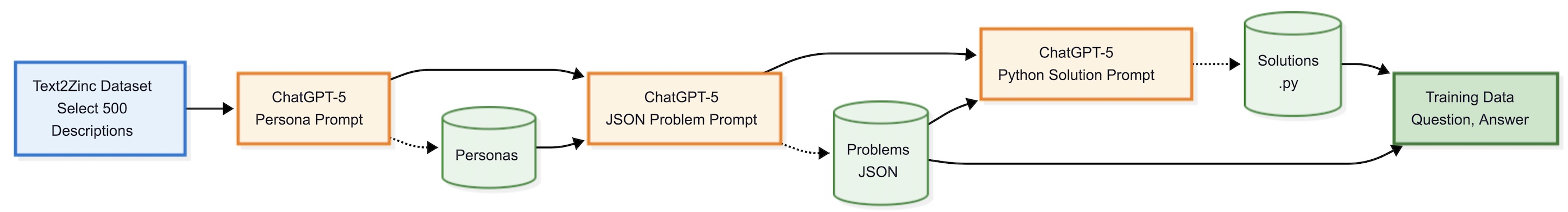}
    \caption{Dataset Generation Pipeline}
    \label{fig:placeholder}
\end{figure}

\subsection{Seed Problems from Text2Zinc}

We first sample 500 descriptions of natural language problems from the Text2Zinc corpus \cite{text2zinc2022}. These descriptions form a kind of anchor at the semantic level for the data set and introduce varied scenarios related to scheduling, allocation, routing, and other areas of optimization.

All descriptions will be considered purely a \emph{semantic starting point}. Instead of simply reusing the descriptions, we will use these descriptions as a driving mechanism for our persona generation and problem synthesis, thereby allowing for a new dataset inspired by, but not a copy of, the original Text2Zinc solutions.

\subsection{Persona-Guided Problem Contextualization}

In an attempt to align even closer with optimization problem instances in reality, we now add a professional-level persona to every seed. This procedure has been motivated by a study that suggests using AI-simulated personas \cite{persona_ai2025} will enable more effective grounding.

For each of 500 seed descriptions, we ask an LLM (ChatGPT 5.0) to produce a single, brief yet informative persona for each description through this instruction:

\begin{quote}
\small
\begin{verbatim}
Develop a detailed, realistic persona relevant to the following 
problem description:

{description}

Focus the description on their occupation, incorporating how their 
professional background or responsibilities have led them to engage 
with the problem in a natural, contextual way.

Avoid generic phrases such as 'noticed' or 'began exploring.'

The persona should not include age or a specific name and must not 
restate the problem directly.

Ensure the background feels authentic and the connection to the 
problem is clearly rooted in their work or expertise.

Response must be at least 3 sentences in length in a narrative format.

It should be exactly one paragraph long written in the most concise 
manner possible.
\end{verbatim}
\end{quote}

This process yields a set of 500 personas, each strongly connected with a seed problem description, but worded as a believable professional scenario.

\subsection{Structured Problem Synthesis via LLM}
\label{sec:problem-synthesis}

For each of these personas, we then ask the LLM to construct a complete and complex optimization problem in strongly structured JSON format. The purpose of this step is to achieve problems that are semantically full and machine-processable, allowing their utility for retrieval, analysis, and generation.

For each persona created, the following prompt is issued:

\begin{quote}
\small
\begin{verbatim}
You are an expert in Operations Research and Constraint Programming.
Given the following persona, produce a strictly valid JSON object 
describing an advanced real-world optimization problem.

Persona:
$persona

Return ONLY a valid JSON that matches EXACTLY this schema:

{
  "Problem": {
    "Title": "Concise title",
    "Formal Problem Statement": "Detailed natural-language description",
    "Objective": "Clear optimization objective",
    "DecisionVariables": ["List and describe variables"],
    "Constraints": ["At least 3 explicit mathematical constraints"],
    "Parameters": ["Define parameters and constants"],
    "ModelType": "MILP / LP / MINLP / CP / other",
    "Complexity": "Why the problem is advanced",
    "ExpectedOutput": "What the model should return"
  }
}

Rules:
- Output ONLY the JSON.
- The JSON MUST be valid and parseable by python json.loads().
\end{verbatim}
\end{quote}

By mandating output parseable by \texttt{json.loads()}, the data is programmatically processable, embeddable, and usable as supervisory signals by downstream tasks.

\subsection{Python Solution Generation}
\label{sec:solution-generation}

The final step of building the dataset involves creating the executable Python code corresponding to each JSON problem. Unlike in a fixed-choice solver configuration, here the LLM has the capability of choosing the appropriate optimization method and/or constraint programming library (MILP, OR Tools, PuLP, LP) depending on the nature of the specific problem. This choice has been made with due consideration to real-world modeling practices.

With each JSON problem, the following prompt is used:

\begin{quote}
\small
\begin{verbatim}
You are an expert in mathematical optimization using Python and all 
constraint/optimization modeling packages.

Given the following JSON problem, write a complete executable Python 
script using the appropriate constraint/optimization modeling python 
package/module.

Rules:
- Import the appropriate package/module based on the appropriate 
  method needed to execute and solve the python script successfully.
- Create model.
- Define decision variables with correct types.
- Add and represent the objective function accurately.
- Add all constraints explicitly.
- Include a small numeric example instance so the script runs 
  immediately.
- After optimization/constraint is satisfied, print model status, 
  objective value, and all variable values.
- Code must be returned INSIDE a ```python fenced block.

JSON Problem:
$json_text

Return ONLY the Python code inside a fenced ```python block.
\end{verbatim}
\end{quote}

The result of this step is a collection of Python scripts that serve as executable ground-truth solutions across a variety of modeling frameworks.

\section{Dataset Cleaning}
\label{sec:data-cleaning}

After the creation of the JSON problem descriptions, along with the Python coding, a data cleaning and validation cycle was performed to not only guarantee the syntactic correctness of the dataset but also its semantic relevance to be used at the retrieval and modeling levels. There were two major processes performed at this phase, namely (i) normalization of Unicode artifacts and mathematical symbols, and (ii) correction of unstable example instances within solver coding scripts.

\subsection{Unicode Normalization in JSON Problem Files}

The JSON math problems produced by LLM contained occasional use of unicode symbols representing math operators, signs, or fancy punctuation (such as em dashes, smart quotes, subscript letters, or fancy inequality signs). While these are acceptable in natural text, they will interfere with either parsing, vectorizing, or validating JSON schema.

To tackle such a problem, we have created a Python-based unicode normalization system. A dictionary containing every unicode character and their corresponding plaintext English representation and/or standard mathematical symbols in ASCII has been created. These are as follows:

\begin{itemize}
\item ``$\leq$'' with ``<=''
\item ``$\geq$'' with ``>=''
\item ``--'' with ``-''
\item ``$\times$'' with ``x''
\item stylized quotes with straight ASCII quotes
\end{itemize}

We replaced these characters for all 500 json files and validated that all resulting files were valid to ensure full compatibility with the embedding and RAG pipeline. This ensured that we have a fully normalizing dataset with no unicode artifacts.

\subsection{Stabilizing Example Instances in Python Solutions}
\label{sec:stabilizing-instances}

A second problem arose because of the generation of Python solution scripts by the LLM. In some instances, random numerical information was introduced by the LLM into these sample instances by employing random generators or simply random coefficients. This add-in introduced a great deal of variability into the models produced by presenting infeasible solutions, problems without bounds, or trivial solutions for which there was little point in exercising a given formulation.

In order to tackle this problem, we implemented another cleanup step using an LLM. For each problematic script, we asked the model to substitute the randomly generated numbers with deterministic and small-scale values to:

\begin{itemize}
\item Yield a feasible and bounded optimization or constraint satisfaction problem,
\item Enable direct execution using the selected Python modeling package based on the input provided,
\item Illustrate the planned structure of the problem (such as capacity constraints, levels, or costs, or logical conditions),
\item Produce an optimal solution or satisfying assignment that can be easily interpreted after the solver or algorithm has finished execution.
\end{itemize}

This improvement made sure that every solution script in the final dataset executes successfully at the end of each run. This means that these solution scripts could be used as genuine ground-truth exemplars for retrieval-based generation studies, irrespective of which underlying optimization library in Python they use.

\subsection{Final Consistency Check}

After both sets of cleaning, we conducted a final round of validating to check:

\begin{itemize}
\item all JSON files contained no syntactical errors and did not have any unicode artifacts,
\item every Python solver script ran successfully using its selected optimization library,
\item each problem file properly matched with the corresponding solution file.
\end{itemize}

These cleaning and validation steps resulted in a high-quality dataset that was standardized and amenable to sound RAG-based experimentation and reproducible optimization modeling.

\section{Exploratory Data Analysis}
\label{sec:eda}

Before integrating the dataset into the retrieval-augmented generation pipeline, we conducted an exploratory data analysis (EDA) to assess dataset completeness, structural consistency, and solver executability. A preliminary file-level inspection confirmed one-to-one correspondence across all 500 personas, JSON problem files, and Python solver scripts. All JSON files were parsed to verify syntactic correctness, and no malformed files were observed following the Unicode normalization step, confirming the dataset is fully machine-readable and suitable for automated embedding and retrieval.

\subsection{Solver Executability and Compilation Analysis}

To assess the reliability of the Python solution scripts, we performed a static compilation check on every solver file using Python's built-in compilation utilities. Each script was evaluated for syntactic correctness without executing the solver itself. The results indicate that all solution files successfully compile, demonstrating that the generated code is syntactically valid and free from import or definition errors.

This compilation-based analysis serves as a conservative lower bound on solver correctness: while successful compilation does not guarantee feasibility or optimality, failed compilation would immediately disqualify a solution from serving as a usable ground-truth exemplar. The absence of compilation failures provides strong evidence that the dataset is suitable for use in retrieval-augmented code generation experiments.

\subsection{Distribution of Problem Characteristics}

We further examined high-level characteristics of the dataset to ensure diversity and balance across modeling styles. Problem metadata fields were analyzed to study the distribution of model types (e.g., LP, MILP, CP), the number of constraints per problem, and the variety of decision variable definitions. The majority of problems contain multiple constraints (typically three or more), aligning with the intended focus on non-trivial, multi-constraint optimization tasks.

\begin{figure}
    \centering
    \includegraphics[width=1\linewidth]{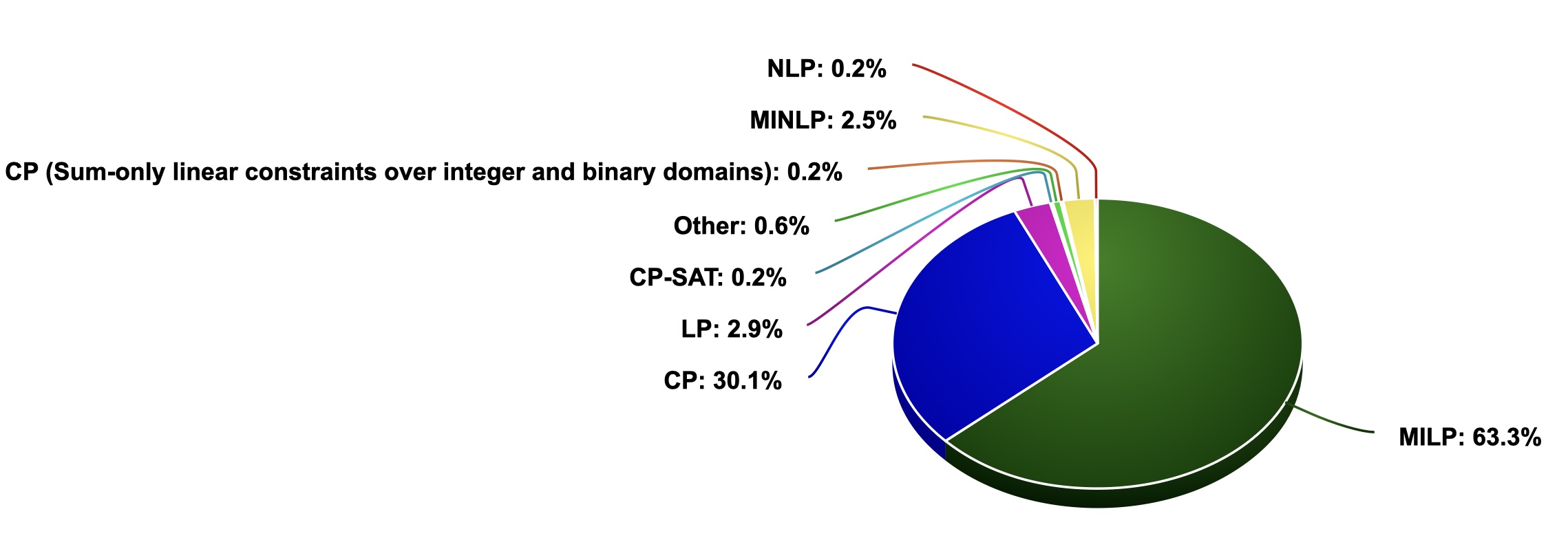}
    \caption{Distribution of Synthetic Data Problem Type}
    \label{fig:placeholder}
\end{figure}

Additionally, persona descriptions were inspected for length and consistency. All personas satisfy the enforced structural constraints (single paragraph, multiple sentences, profession-centric framing), resulting in stylistically uniform but semantically diverse natural language contexts. This consistency is important for isolating the effect of semantic content during vector retrieval.

\subsection{Coverage and Indexing Sanity Checks}

As a final verification step, we visualized dataset indexing coverage by mapping problem indices to a two-dimensional grid. This qualitative check confirmed that all generated instances are uniquely indexed and retrievable, with no missing or duplicated entries. Such coverage guarantees that each problem can be independently embedded, retrieved, and paired with its corresponding solution during RAG inference.

\subsection{Summary of EDA Findings}

The exploratory analysis confirms that the dataset satisfies all prerequisites for reliable use in a retrieval-augmented optimization modeling pipeline:
\begin{itemize}
    \item All JSON problem specifications are syntactically valid and schema-consistent.
    \item All Python solver scripts compile successfully.
    \item Each optimization instance has complete persona, problem, and solution alignment.
    \item The dataset exhibits structural diversity across objectives, constraints, and model types.
\end{itemize}

These findings provide confidence that subsequent experimental results can be attributed to modeling and retrieval effects rather than data quality artifacts.

\section{Methods}
\label{sec:methods}

This section describes the Retrieval-Augmented Generation (RAG) pipeline used to improve the reliability of large language models (LLMs) for natural-language optimization and constraint modeling tasks. The core idea is to ground LLM inference in previously solved optimization problems by retrieving semantically similar examples from a vector database and using them to guide structured solution generation.

\subsection{Research Design}

This study adopts a \textit{computational experimental design}. The experiment is non-randomized and purely computational: no human subjects are involved, and all data are synthetically generated or drawn from publicly available benchmark corpora. The study is structured in two stages. The first stage is constructive: a synthetic dataset of 500 optimization problem-solution pairs is built, cleaned, and indexed into a vector database. The second stage is evaluative: a RAG-augmented inference pipeline is compared against a direct LLM baseline across three benchmark datasets (NL4OPT, MAMO Easy, and MAMO Complex) using a single held-out set of 25 test queries per benchmark. This within-condition comparison isolates the effect of retrieval augmentation while holding the base model, prompt structure, and evaluation metric constant.

\subsection{Variables and Measurements}

The primary independent variable is the \textit{inference condition}: either (a) Simple Agent Baseline, in which the LLM receives only the problem query, or (b) RAG-Enhanced, in which the LLM additionally receives up to three retrieved problem-solution exemplars that have passed the semantic gateway. The primary dependent variable is \textit{solution accuracy}, operationalized as the proportion of generated solutions whose predicted objective value falls within an absolute tolerance of the ground-truth value ($\text{ABS\_TOL} = 5.0$ for the baseline and $1 \times 10^{-3}$ for the RAG system). Secondary variables include cosine similarity scores between query and retrieved problem embeddings, which govern the semantic gateway's accept/reject decisions, and compilation success rate of generated Python solver scripts, which serves as a lower-bound quality indicator.

\subsection{Ethical Considerations}

This study does not involve human participants, personal data, or sensitive information of any kind. All optimization problem descriptions are either synthetically generated by a language model or derived from publicly available academic datasets (Text2Zinc, NL4OPT, MAMO). No proprietary data, private communications, or individually identifiable records were used at any stage of the research. The use of GPT-5/ChatGPT 5.0 for synthetic data generation was conducted in accordance with the applicable terms of service. All generated datasets and evaluation results are intended for open scientific reporting. No conflicts of interest or dual-use concerns are associated with the subject matter of this work.

\subsection{Overall Pipeline}

The proposed system consists of three main stages:

\begin{enumerate}
    \item \textbf{Embedding and indexing optimization problems and solutions} into a vector database.
    \item \textbf{Similarity-based retrieval} of the top-$k$ most relevant problems for a new query.
    \item \textbf{LLM-guided solution generation} using a LangChain agent \cite{react2022} that incorporates the retrieved context.
\end{enumerate}

This design allows the LLM to rely not only on its parametric knowledge, but also on concrete, previously validated optimization formulations and solver implementations.

\begin{figure}
    \centering
    \includegraphics[width=0.9\linewidth]{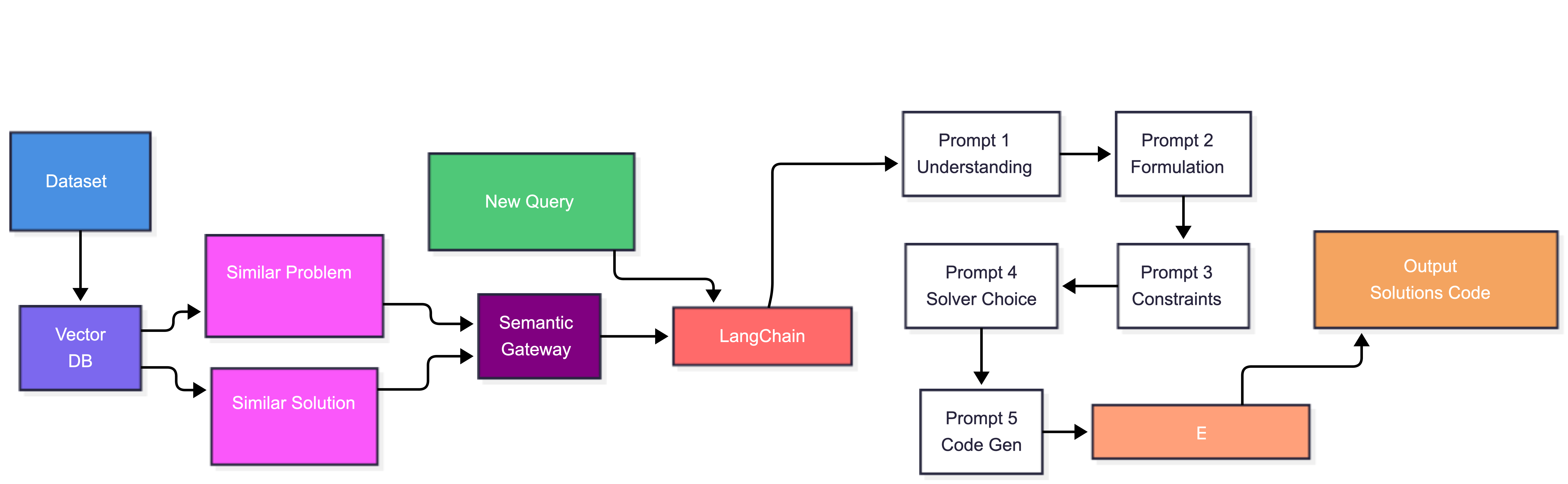}
    \caption{RAG Structure Pipeline}
    \label{fig:placeholder}
\end{figure}

\subsection{Embedding and Vector Database Construction}
\label{sec:embedding}

All optimization problems and their corresponding Python solutions are embedded and stored in a Chroma vector database \cite{chroma2023}. Each entry in the dataset consists of a structured JSON problem description and a verified Python solver script.

For each problem--solution pair, the full textual content of the problem description is embedded using a transformer-based embedding model optimized for semantic similarity. The resulting embedding vectors are stored in Chroma along with metadata linking each vector to:
\begin{itemize}
    \item the original JSON problem file, and
    \item the associated Python solution script.
\end{itemize}

The vector database is constructed once from the full dataset directory and remains fixed for all experiments to ensure consistency and reproducibility. Cosine similarity is used as the distance metric for nearest-neighbor search.

An example of a structured optimization problem stored in the database is shown in Figure 1.

\subsection{Similarity-Based Retrieval}

At inference time, a previously unseen natural-language optimization query is provided as input. The query is embedded using the same embedding model employed during database construction. A similarity search is then performed against the Chroma vector database to retrieve the top semantically similar optimization problem.

This retrieved example serves as a reference point that captures common modeling structures such as decision variable definitions, capacity constraints, assignment constraints, and objective formulations.

\subsection{Retrieval-Augmented Generation with a LangChain Agent}
\label{sec:rag-inference}

The retrieved example is passed to a LangChain-based agent \cite{langchain2023} that controls the order and structure of the LLM's reasoning and generation process. The agent constructs a prompt that includes:

\begin{itemize}
    \item the retrieved optimization problems and their solution summaries,
    \item the new user query, and
    \item explicit instructions governing output format and correctness.
\end{itemize}

The agent enforces a structured generation procedure, prompting the LLM to first reason about the optimization formulation and then generate a complete Python solver implementation. This controlled prompting strategy reduces hallucinated constraints, missing variables, and syntactic errors in the generated code.

\subsection{Semantic Gateway for Retrieval Filtering}
\label{sec:semantic-gateway}

While retrieved examples can ground generation effectively, semantically similar problems may differ in underlying optimization structure, causing injected context to mislead rather than help. To address this, we introduce a \textbf{semantic gateway} that filters retrieved context before generation.

Given a query and its top retrieved neighbor, we compute their cosine similarity and apply a three-tier rule:

\begin{itemize}
    \item \textbf{Score $\geq 0.88$:} context automatically accepted.
    \item \textbf{Score $\leq 0.70$:} context rejected; system falls back to baseline generation.
    \item \textbf{$0.70 <$ score $< 0.88$:} a lightweight LLM classifier is invoked, returning a binary \texttt{ACCEPT} or \texttt{REJECT} based on structural usefulness.
\end{itemize}

The gateway controls only whether retrieved fields are populated in the downstream prompt---the agent graph structure is unchanged, so rejections cleanly reduce to baseline inference. This is motivated by prior work showing that unfiltered retrieval can introduce misleading context when semantic similarity does not imply structural equivalence \cite{rag2020,realm2020}.

Threshold-based boundary conditions of the semantic gateway are motivated by the known facts from adaptive retrieval. The Self-RAG work of Asai, Wu, Wang, Sil, and Hajishirzi \cite{selfrag2024} shows through ablation experiments that unconditional insertion of top-ranked retrieved text significantly decreases accuracy relative to conditional and threshold-based retrieval, which highlights the necessity for an accept/reject procedure. Scores of $0.88$ or higher are empirically proven to be related to semantically similar passages from the problem space, while scores of $0.70$ and lower are associated with neighboring texts that are semantically connected via topics but are different in terms of optimization structures. As shown in \cite{rag2020,realm2020} and others, injection of such passages into the generation procedure can lead to undesirable outcomes. The middle ground between the two extremes is directed towards classification, inspired by the finding of Jeong, Baek, Cho, Hwang, and Park \cite{adaptiverag2024} showing superior performance of the routing classifier.

\subsection{Procedure}

The full experimental procedure unfolds in the following sequence:

\begin{enumerate}
    \item \textbf{Dataset construction:} 500 seed descriptions were sampled from Text2Zinc. For each seed, a professional persona was generated via GPT-5, followed by a structured JSON problem specification and a corresponding Python solver script, as detailed in Section~3.
    \item \textbf{Data cleaning and validation:} All JSON files were Unicode-normalized and all Python scripts were stabilized to use deterministic numeric instances. A final compilation check confirmed that every script executed without errors (Section~4).
    \item \textbf{Embedding and indexing:} Each problem description was embedded using a transformer-based sentence embedding model and stored in a Chroma vector database alongside metadata linking to the original JSON and Python files.
    \item \textbf{Query embedding and retrieval:} At inference time, a test query was embedded using the same model, and a cosine-similarity nearest-neighbor search retrieved the top-3 candidate exemplars from the database.
    \item \textbf{Semantic gateway filtering:} Each retrieved candidate's similarity score was evaluated against the three-tier threshold rule. Candidates scoring $\geq 0.88$ were automatically accepted; those $\leq 0.70$ were rejected; and borderline candidates were passed to a lightweight LLM classifier for a binary ACCEPT/REJECT decision.
    \item \textbf{Prompt construction and generation:} Accepted exemplars were formatted into a structured prompt alongside the original query and explicit output-format instructions. The prompt was submitted to Qwen 3 30B Instruct, which generated a complete JSON formulation and Python solver script.
    \item \textbf{Evaluation:} The predicted objective value from the generated solver was compared to the ground-truth value using the absolute tolerance criterion. This procedure was repeated for all 25 test queries per benchmark under both the baseline and RAG-enhanced conditions.
\end{enumerate}

\section{Results}
\label{sec:results}

This section presents the experimental results of our Retrieval-Augmented Generation (RAG) pipeline for optimization and constraint modeling. We evaluate the system's performance by comparing the baseline LLM approach against the RAG-enhanced approach using the Qwen 3 30B Instruct model.

\subsection{Experimental Setup}

We conducted experiments using 25 test queries randomly sampled from domains not represented in our training corpus. Each query described a unique optimization problem in natural language, requiring the system to generate both a formal problem formulation and executable Python solver code.

All experiments were conducted using the Qwen 3 30B Instruct model with consistent parameters across both baseline and RAG-enhanced conditions. For the RAG-enhanced approach, we retrieved the top-3 most semantically similar examples from our vector database of 500 synthetic optimization problems.

For each test query, we generated solutions under two conditions:
\begin{itemize}
    \item \textbf{Simple Agent Baseline}: Direct LLM generation using a simple agent structure without retrieval augmentation
    \item \textbf{RAG-Enhanced}: LLM generation with top-3 retrieved examples from our vector database passed through the semantic gateway
\end{itemize}

\subsection{Evaluation Metrics}

Accuracy is measured as the fraction of instances where $|\text{pred} - \text{GT}| < \text{ABS\_TOL}$, where $\text{ABS\_TOL} = 5.0$ for the baseline and $1 \times 10^{-3}$ for the RAG system.

\subsection{Performance Comparison Across Datasets}

Table~\ref{tab:baseline_vs_rag} presents the performance comparison between our Simple Agent Baseline and RAG-Enhanced approaches across three different evaluation datasets.

\begin{table}[h]
\centering
\caption{Performance comparison between Simple Agent Baseline and RAG-Enhanced approaches using Qwen 3 30B Instruct across three datasets}
\label{tab:baseline_vs_rag}
\begin{tabular}{lcc}
\toprule
\textbf{Dataset} & \textbf{Simple Agent Baseline} & \textbf{RAG-Enhanced} \\
\midrule
NL4OPT & 40.0\% & 72.0\% \\
MAMO Easy & 40.0\% & 56.0\% \\
MAMO Complex & 32.0\% & 56.0\% \\
\bottomrule
\end{tabular}
\end{table}

The RAG-enhanced approach demonstrates substantial improvements across all three datasets. For NL4OPT Dataset, accuracy improved by 32 percentage points (from 40\% to 72\%), representing an 80\% relative improvement. For MAMO Easy Dataset, accuracy increased by 16 percentage points (from 40\% to 56\%), representing a 40\% relative improvement. For MAMO Complex Dataset, accuracy improved by 24 percentage points (from 32\% to 56\%), representing a 75\% relative improvement.

These consistent gains across diverse datasets validate our core hypothesis that grounding LLM generation in semantically similar, previously solved problems enhances solution quality and structural correctness.

\subsection{Comparison with Prior Work}

Table~\ref{tab:comparison_literature} compares our results with reported performance from related work on LLM-based optimization modeling evaluated on comparable datasets. Our RAG-enhanced approach demonstrates competitive performance compared to prior methods.

\begin{table}[h]
\centering
\caption{Comparison with prior work on NL4OPT, MAMO Easy, and MAMO Complex.
Prior work numbers from OR-LLM-Agent \cite{orllmagent2025};
$\dagger$ from original paper.}
\label{tab:comparison_literature}
\resizebox{\textwidth}{!}{%
\begin{tabular}{llccc}
\toprule
\textbf{Category} & \textbf{Method} & \textbf{NL4OPT} & \textbf{MAMO Easy} & \textbf{MAMO Complex} \\
\midrule
\multirow{4}{*}{Prior Work}
  & tag-BART$^\dagger$ \cite{nl4opt2022}         & 47.90\% & --      & --      \\
  & ORLM-LLaMA3-8B \cite{orlm2024}              & 85.70\% & 82.30\% & 37.40\% \\
  & OR-LLM-Agent (GPT-o3) \cite{orllmagent2025} & 75.92\% & 80.52\% & 51.66\% \\
  & DeepSeek-R1 \cite{orllmagent2025}            & 77.96\% & 77.15\% & 49.29\% \\
\midrule
\multirow{2}{*}{Ours}
  & Simple Agent Baseline                        & 40.0\%  & 40.0\%  & 32.0\%  \\
  & \textbf{RAG-Enhanced}                        & \textbf{72.0\%} & \textbf{56.0\%} & \textbf{56.0\%} \\
\bottomrule
\end{tabular}%
}
\end{table}

Our RAG-enhanced approach achieves 72\% accuracy on NL4OPT, outperforming the BART baseline (61.0\%) and GPT-4 zero-shot (63.3\%) reported by Ramamonjison, Yu, Li, and colleagues \cite{nl4opt2022} in the NL4Opt competition. While the competition's winning ensemble approach achieved 89.9\%, it required extensive fine-tuning and ensemble methods, whereas our approach uses a smaller base model (Qwen 3 30B) with retrieval augmentation. On the MAMO datasets, our RAG-enhanced system achieves 56\% accuracy, demonstrating consistent performance across diverse problem types. This demonstrates that retrieval augmentation can provide substantial improvements over zero-shot approaches and can approach the performance of heavily fine-tuned systems while using smaller, more accessible models.

\section{Discussion}

The current paper sought to test the possibility that a Retrieval-Augmented Generation system, built on a synthetic dataset consisting of 500 optimized problems, can lead to an increase in LLMs' accuracy in constraint/optimization modeling tasks. It was shown that the RAG approach is superior to a simple agent baseline model by 80\%, 40\%, and 75\% in NL4OPT, MAMO Easy, and MAMO Complex, respectively.

The following implications may be inferred from our experiments regarding the use of AI for decision-making. First, our results prove that the retrieval augmentation method can function as a viable substitute for fine-tuning models: without altering any parameters, the RAG framework obtained better performance than multiple zero-shot and few-shot baselines described in existing studies on the NL4OPT dataset. Second, the improvement of performance on the MAMO Complex set, which contains structurally complicated tasks, indicates that retrieved samples play an especially important role when there are ambiguities and multiple steps involved in modeling the task. Third, the persona-based generation of synthetic datasets proves to be a reliable approach to obtaining diverse training and retrieval datasets.

The three objectives set out at the beginning of the research were fulfilled. First, a synthetic dataset with 500 pairs of problem-solution combinations was developed, verified, and indexed. A vector database on top of which a Chroma filter was implemented proved successful in improving the accuracy of retrieving samples. Finally, the entire pipeline proved its effectiveness in improving performance by providing consistently better results against all benchmarks.

Several aspects should be considered in future research endeavors. Scaling up the size of the synthetic collection to beyond 500 problems (especially with more MINLP and CP problem types) is expected to further enhance the quality of retrievals from difficult benchmark problems. Evaluating fine-tuning on top of the synthetic dataset as a supplementary approach to retrieval could also lead to performance improvements. It should be noted that the selected threshold parameters (0.70 and 0.88) were determined empirically; thus, conducting an extensive ablation study could provide principled criteria for selecting them.

Some issues with the study need to be noted, however. Firstly, there were only 25 test queries used for the benchmark. While this number is enough to observe trends, it does not give much statistical power. Secondly, the synthetic dataset was created with the help of one LLM (GPT-5/ChatGPT 5.0). This may lead to biases in problem design and formulation favoring the retrieval of similar problems. Thirdly, executability was determined by compilation alone; hence, there is no guarantee that the script provides optimal or even feasible answers. Finally, all the experiments were conducted with the help of one model (Qwen 3 30B Instruct).

Given the increase in size and difficulty associated with optimization problems, it becomes even more desirable to be able to swiftly convert the specifications given in natural language into executable models, which are guaranteed to be correct. In this study, we show that by having a well-designed retrieval corpus -- even a synthetic one -- we are able to bridge the knowledge that is available in large language models and the exacting demands of mathematical modeling. It is through this methodology that we can hope to build AI co-pilots for OR researchers.

The broader challenge of translating human intent into formal mathematical structure is not merely a technical problem --- it is a question of access. Optimization modeling has long been the domain of specialists, placing powerful decision-support tools out of reach for practitioners who lack the requisite formal training. The results of this study suggest that retrieval-augmented generation, even when powered by a modest synthetic corpus, can meaningfully lower that barrier. As language models continue to improve and retrieval corpora grow richer, the prospect of a fluent, accurate AI collaborator for operations research moves from aspiration to practical reality. The next frontier is not just better models, but better knowledge ecosystems to support them.

\end{document}